%% file: root.tex
\documentclass[letterpaper, 10 pt, conference]{ieeeconf}
\IEEEoverridecommandlockouts
\usepackage{hyperref}
\usepackage{amssymb}
\usepackage{amsmath}
\usepackage{graphicx}
\usepackage{url}
\usepackage{multirow}
\usepackage{xcolor,colortbl}
\usepackage[ruled,linesnumbered, noend]{algorithm2e}
{\end{list}}

\title{\LARGE \bf Improving 6D Pose Estimation of Objects in Clutter\\ via
  Physics-aware Monte Carlo Tree Search}
\author{Chaitanya Mitash, Abdeslam Boularias and Kostas E. Bekris% <-this % stops a space
\thanks{The authors are with the Computer Science Department of
  Rutgers University in Piscataway, New Jersey, 08854, USA. Email:
  \{cm1074,kb572,ab1544\}@rutgers.edu.}
\thanks{This work is supported by NSF awards IIS-1734492, IIS-1723869
  and IIS-1451737. Any opinions or findings expressed in this paper do
  not necessarily reflect the views of the sponsors.}%
}

\begin{document}
\maketitle
\thispagestyle{empty}
\pagestyle{empty}

\begin{abstract}
\input{00_abstract}
\end{abstract}

\section{Introduction}
\label{sec:intro}
\input{01_introduction}

\section{Related Work}
\label{sec:related}
\input{02_relatedwork}

\section{Approach}
\label{sec:approach}
\input{03_approach}

\section{Evaluation}
\label{sec:evaluation}
\input{04_evaluation}

\section{Discussion}
\label{sec:conclusion}
\input{05_conclusion}

\bibliographystyle{IEEEtran}
\bibliography{root}

\end{document}

%% file: 00_abstract.tex
This work proposes a process for efficiently searching over
combinations of individual object 6D pose hypotheses in cluttered
scenes, especially in cases involving occlusions and objects resting
on each other. The initial set of candidate object poses is generated
from state-of-the-art object detection and global point cloud
registration techniques. The best scored pose per object by using
these techniques may not be accurate due to overlaps and occlusions.
Nevertheless, experimental indications provided in this work show that
object poses with lower ranks may be closer to the real poses than
ones with high ranks according to registration techniques. This
motivates a global optimization process for improving these poses by
taking into account scene-level physical interactions between
objects. It also implies that the Cartesian product of candidate poses
for interacting objects must be searched so as to identify the best
scene-level hypothesis. To perform the search efficiently, the
candidate poses for each object are clustered so as to reduce their
number but still keep a sufficient diversity. Then, searching over the
combinations of candidate object poses is performed through a Monte
Carlo Tree Search (MCTS) process that uses the similarity between the
observed depth image of the scene and a rendering of the scene
given the hypothesized pose as a score that guides the search
procedure. MCTS handles in a principled way the tradeoff between
fine-tuning the most promising poses and exploring new ones, by using
the Upper Confidence Bound (UCB) technique. Experimental results
indicate that this process is able to quickly identify in cluttered
scenes physically-consistent object poses that are significantly
closer to ground truth compared to poses found by point cloud
registration methods.

%% file: 01_introduction.tex
\begin{figure}[t]
\centering \includegraphics[width=\linewidth]{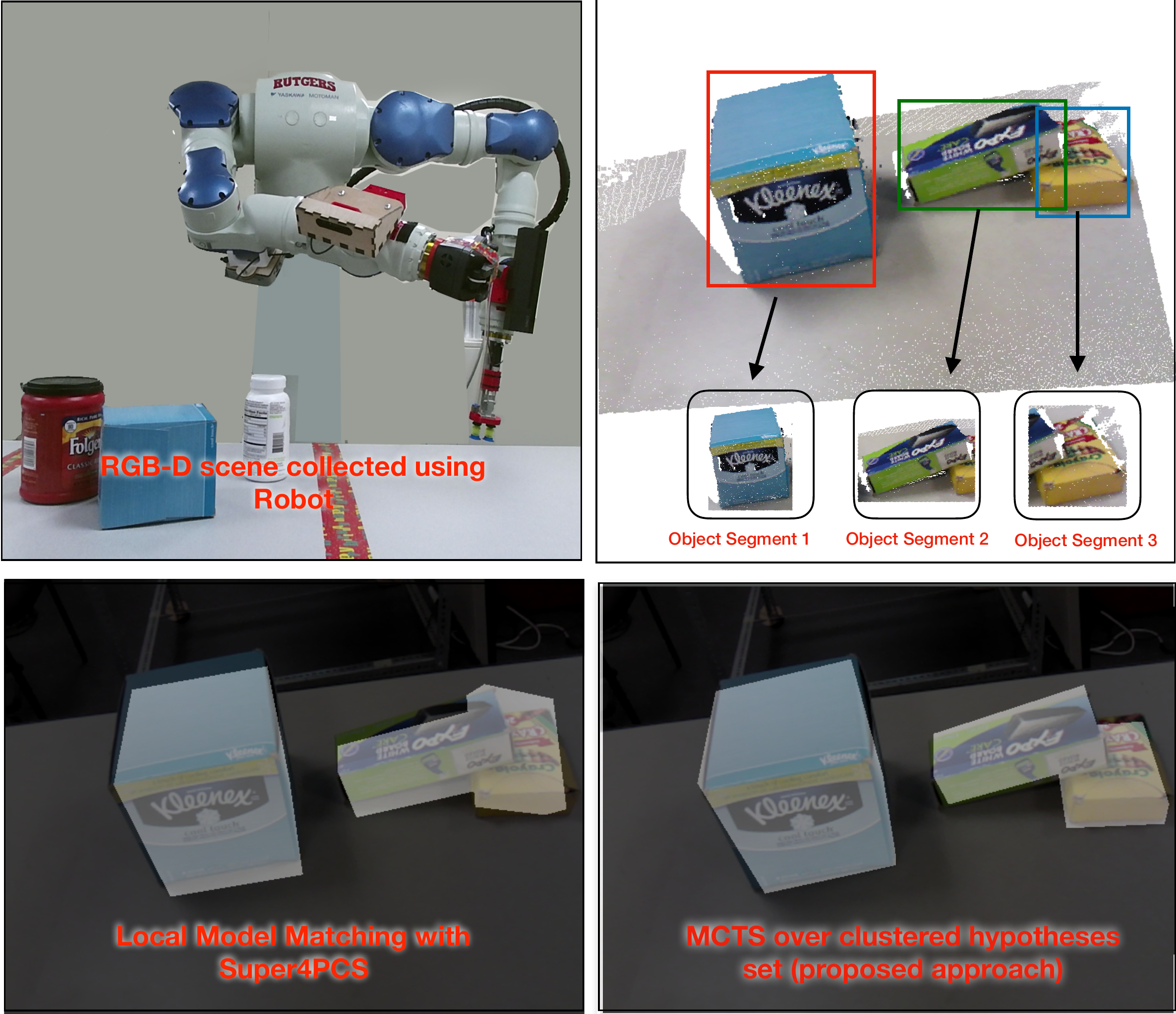}
\vspace{-.25in}
\caption{(top left) A Motoman robot using Intel RealSense to capture RGB-D image of the scene. (top-right) An example scene with 3 objects and resulting segmentation achieved with state-of-the-art
method \cite{ren2015faster}. (bottom left) The pose estimation according to the best hypothesis from
Super4PCS \cite{super4pcs}. Notice the penetration between the two objects on the right side of the scene. (bottom right) Improved pose estimation via the proposed physics-aware search process, which
considers the ensemble of hypotheses returned by Super4PCS.}
\label{fig:intro}
\vspace{-.25in}
\end{figure}

Robot manipulation systems frequently depend on a perception pipeline that can accurately perform object recognition and six degrees-of-freedom (6-{\tt DOF}) pose estimation. This is becoming increasingly important as robots are deployed in less structured environments than traditional automation setups. An example application domain corresponds to warehouse automation and logistics,
as highlighted by the Amazon Robotics Challenge ({\tt ARC}) \cite{Correll:2016aa}. In such tasks, robots have to deal with a large variety of objects potentially placed in complex arrangements, including cluttered scenes where objects are resting on top of each other and are often only partially visible, as shown in Fig. \ref{fig:intro}.

This work considers a similar setup, where a perception system has
access to {\tt RGB-D} images of objects in clutter, as well as {\tt 3D
CAD} models of the objects, and must provide accurate pose estimation
for the entire scene. In this domain, solutions have been developed
that use a Convolutional Neural Network ({\tt CNN}) for object
segmentation \cite{Princeton, Hernandez:2016aa} followed by a {\tt 3D}
model alignment step using point cloud registration techniques for
pose estimation \cite{super4pcs, icp}. The focus of the current paper
is to improve this last step and increase the accuracy of pose
estimation by reasoning at a scene-level about the physical
interactions between objects.

In particular, existing object-level reasoning for pose estimation can fail on several instances. One reason can be imperfect object detection, which might include parts of other objects, thus guiding
the model registration process to optimize for maximum overlap with a mis-segmented object point cloud. Another challenge is the ambiguity that arises when the object is only partially visible, due to occlusion. This results in multiple model placements with similar alignment scores, and thus a failure in producing a unique, accurate pose estimate. These
issues were the primary motivation behind hypothesis verification
({\tt HV}) methods \cite{aldoma2016global, aldoma2013multimodal}. These techniques follow a pipeline where: (a) they first generate multiple candidate poses per object given feature matching against a model, (b) they create a set of scene-level hypotheses by considering the Cartesian product of individual object pose candidates, and find the optimal hypothesis with respect to a score defined in terms of similarity with the input scene and geometric constraints.

It has been argued, however, that existing {\tt HV} methods suffer from critical limitations for pose estimation \cite{Narayanan:2016aa, d2p}. The argument is that the optimization in the {\tt HV} process, may not work because the true poses of the objects may not be included in the set of generated candidates, due to errors in the process for generating these hypotheses. Errors may arise from the fact that the training for detection typically takes place for individual objects and is not able to handle occlusions in the case of cluttered scenes.

This motivated the development of a search method \cite{Narayanan:2016aa, d2p} for the best explanation of the
observed scene by performing an exhaustive but informed search through
rendering all possible scene configurations given a discretization
over {\tt 3-DOFs}, namely $(x, y, yaw)$. The search was formulated as
a tree, where nodes corresponded to a subset of objects placed at
certain configurations and edges corresponded to adding
one more object in the scene. The edge cost was computed based on the
similarity of the input image and the rendered scene for the the additional object. An order of object
placements over the tree depth was implicitly defined by constraining
the child state to never occlude any object in the parent state. This
ensured an additive cost as the search progressed, which allows the
use of heuristic search.

\begin{figure*}[h]
\centering \includegraphics[width=\textwidth]{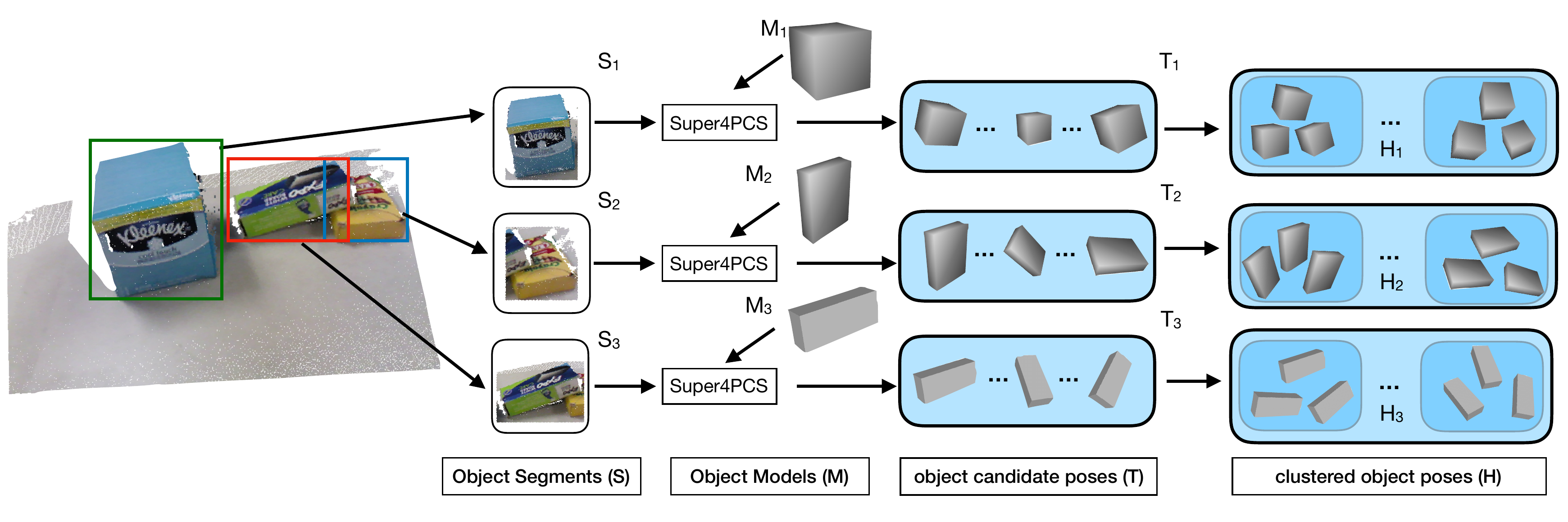}
\vspace{-.25in}
\caption{The image describes the process of hypotheses generation for
objects present in the scene. The process starts with extracting
object segments S$_{1:3}$ using Faster-RCNN \cite{ren2015faster},
followed by using a global point cloud registration
technique \cite{super4pcs} to compute a set of possible model
transformations (T$_{1:3}$) that corresponds to the respective
segments. These transformations are then clustered to produce object
specific hypotheses sets (H$_{1:3}$).} \vspace{-.2in}
\label{fig:hypogen}
\end{figure*}

The current work adapts this idea of tree search to achieve better
scalability and increased accuracy, while performing a comprehensive
hypothesis verification process by addressing the limitations of such
{\tt HV} techniques. In particular, instead of imposing a discretization, which is
difficult to scale to {\tt 6-DOF} problems, the proposed search is
performed over scene hypotheses. In order to address the issue of
potentially conflicting candidate object poses, the scene hypotheses
are dynamically constructed by introducing a constrained local
optimization step over candidate object poses returned by Super4PCS, a
fast global model matching method \cite{super4pcs}. To limit
detection errors that arise in cluttered scene, the proposed method
builds on top of a previous contribution \cite{Mitash:2017aa}, which
performs clutter-specific autonomous training to get object
segments. This paper provides experimental indications that the set of
candidate object poses returned by Super4PCS given the clutter-aware
training contains object poses that are close enough to the ground truth, however,
these might not be the ones that receive the best matching score according to
Super4PCS. This is why it becomes necessary to search over the set of candidate poses. Searching over all possible hypotheses returned by Super4PCS, however,
is impractical. Thus, this work introduces a clustering approach that identifies a small set of
candidate pose representatives that is also diverse enough to express the spread of guesses in the matching process.
 
The search operates by picking candidate object poses from the set of cluster representatives given a specific order of object placement. This order is defined by
considering the dependencies amongst objects, such as, when an object
is stacked on top of another or is occluded by another
object. At every expansion of a new node, the method uses the previously placed objects to re-segment the object point cloud. It then performs local point cloud registration as
well as physics simulation to place the object in physically
consistent poses with respect to the already placed objects and the
resting surface. As the ordering considers the physical dependency
between objects, the rendering cost is no longer additive and cannot
be defined for intermediate nodes of the search tree. For this reason,
a Monte Carlo Tree Search ({\tt MCTS})~\cite{kocsis2006bandit}
approach is used to heuristically guide the search based on evaluation
of the rendering cost for the complete assignment of object
poses. Specifically, the UCT search method is used with the Upper
Confidence Bound (UCB) to trade-off exploration and exploitation in the
search tree.

The experimental evaluation of the proposed framework indicates that
searching over the space of scene hypotheses in this manner can
quickly identify physically realistic poses that are close to ground
truth and can significantly improve the quality of the output of
global model matching methods.

%% file: 02_relatedwork.tex
This section covers the different approaches that have been applied to
object recognition and pose estimation using range data and their
relationship to the current work.

\subsection{Feature Matching}

A traditional way of approaching this problem has been computing local shape based descriptors in the observed scene and on the {\tt 3D CAD} object models \cite{aldoma2012tutorial}. Given feature correspondence between the real scene and the model, rigid body transformations can be computed for the models that maximize the correspondences between these descriptors. Some examples of such local
descriptors correspond to the {\tt SHOT} \cite{salti2014shot}, and {\tt PFH} \cite{rusu2008aligning}. There has also been significant use of viewpoint-specific global shape features like {\tt CVFH} \cite{aldoma2011cad}. In this case, features are computed offline for object models viewed from several sampled viewpoints and are matched to the feature
corresponding to the object segment retrieved in the observed data to
directly get the pose estimate. These methods have gained popularity
because of their speed of execution and minimal training requirement.
However, the local descriptors are prone to failure in
case of occlusion when the keypoints are not visible and the global
descriptors are highly dependent on clean object segmentation, which is often difficult to obtain.

\subsection{Progress in Deep Learning}

There has been recent success in using deep learning techniques for
learning local geometric descriptors \cite{zeng20163dmatch} and local
{\tt RGB-D} patches \cite{kehl2016deep} for object recognition and
pose estimation. Eventhough this is promising as data driven
approaches could help close the gap between 3d geometric models and
noisy observed data, it needs access to a large model aligned 3d
training dataset, which may be difficult to collect. Another technique
often used is to perform object segmentation using {\tt CNNs} trained
specifically for the setup \cite{Mitash:2017aa, Princeton,
schwarz2016rgb} and perform point cloud registration methods
\cite{super4pcs, icp} to align 3d models to the object segments. This
method may fail in the case of over-segmentation or when enough
surfaces are not visible.

\subsection{Global Scene-Level Reasoning}

A popular approach to resolve conflicts arising from local reasoning is to generate object pose candidates, and perform a Hypothesis Verification ({\tt HV}) step \cite{aldoma2012global, aldoma2013multimodal, akizuki2016physical}. The hypotheses generation in most cases occurs using a variant of {\tt RANSAC} \cite{RANSAC, super4pcs}. One of this method's drawbacks is that the generated hypotheses might already be conflicted due to errors in object segmentation and thus
performing an optimization over this might not be very useful. Recently, proposed method reasons globally about hypotheses generation process \cite{michel2016global}. Nevertheless, this requires explicit training for pixelwise prediction. Another approach to counter these drawbacks corresponds to an
exhaustive but informed search to find the best scene hypotheses over a discrete set of object placement configurations \cite{Narayanan:2016aa, d2p}. A tree search formulation as described above was defined to effectively search in {\tt 3-DOF} space. It is not easy, however, to apply the method for {\tt 6-DOF} pose estimation due to scalability issues.

This work shows that by training an object detector with an autonomous
clutter-aware process \cite{Mitash:2017aa}, it is possible to generate
a set of object candidate poses by a fast global point cloud
registration method \cite{super4pcs}, which only has local geometric
conflicts. Generating candidate poses in this manner and then applying
a search process, which constrains each object expansion to other
object placements leads to significant improvements in the final pose
estimation results.

%% file: 03_approach.tex
The problem involves estimating {\tt 6-DOF} poses of objects placed in
a clutter, the inputs to which are:
\begin{itemize}
\item {\tt RGB} and {\tt D} images captured by a sensor;
\item a set of 3D CAD models {\tt M}, one for each object.
\item a known {\tt 6-DOF} pose of the camera: {\tt cam\_pos}.
\end{itemize}

The proposed method approaches the problem by (1) generating a set of
pose hypotheses for each object present in the scene, and (2)
searching efficiently over the set of joint hypotheses for the most
globally consistent solution. Global consistency is quantitatively
evaluated by a score function. The score function measures the
similarity between the actual observed depth image and a rendering of
the objects in simulation using their hypothesized poses.  The
hypothesized poses are adapted during the search process, so as to
correspond to poses where the objects are placed in a physically
realistic and stable configuration according to a physics engine that
simulates rigid object dynamics.

\subsection{Hypothesis Generation}

Some of the desired properties for a set of 6D pose hypotheses are the
following:

\begin{itemize}
\item informed and diverse enough such that the optimal solution is either already
  contained in the set or a close enough hypothesis exists so that a
  local optimization process can fine-tune it and return a good
  result;
\item limited in size, as evaluating the dependencies amongst the
hypotheses set for different objects can lead to a combinatorial
explosion of possible joint poses and significantly impact the
computational efficiency;
\item does not require extensive training.
\end{itemize}

This work aims to consider all of these properties while generating
the hypothesis set. The pseudocode for hypothesis generation is presented in
Algorithm~\ref{alg:alg1}.

\vspace{-.1in}
\input{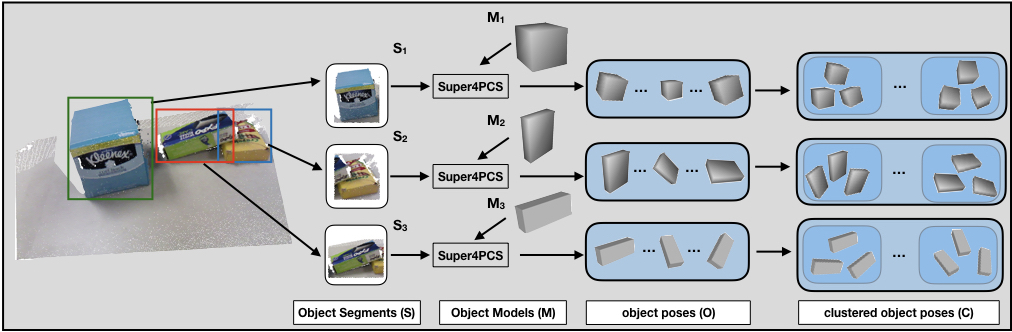}
\vspace{-.1in}

The algorithm first employs an object detector on the {\tt RGB} image
based on {\tt Faster-RCNN} \cite{ren2015faster}. This detector, which
uses a {\tt VGG16} network architecture \cite{simonyan2014very}, was
trained with an autonomous training process proposed in prior
work \cite{Mitash:2017aa}. The training involves a large number of
realistic synthetic data used to bootstrap a self-learning process to
get a high performing detector. For each object, the corresponding
bounding-box (bbox) returned by the object detector is used to get a
segment $P_o$ of the 3D point cloud. Segment $P_o$ is a subset of the
point cloud of the scene and contains points from the visible part of
object {\tt o}. Segment $P_o$ frequently contains some points from
nearby objects because the bounding box does not perfectly match the
shape of the object.

The received point set $P_o$ is then matched to the object model {\tt
M$_o$} by using the {\it Super4PCS}
algorithm \cite{super4pcs}. Super4PCS finds sets of congruent 4-points
in the two point clouds related by a rigid transformation and returns
the transformation, which results in the best alignment according to
the {\tt LCP} (Largest Common Pointset) measure. Nevertheless, this
returned transformation is not necessarily the optimal pose of the
object as the point cloud segment extracted via the detection process
could include parts of other objects or due to lack of visible surface
might not be informative enough to compute the correct solution.

This motivates the consideration of other possible transformations for
the objects, which can be evaluated in terms of scene-level
consistency. This is the focus of the following section. Thus, {\sc
Super4PCS} is used to generate a set of possible transformations {\tt
T$_o$} computed using congruent 4-point sets within a time budget {\tt
t$_o$}. It is interesting to consider the quality of the hypotheses
set returned by the above process by measuring the error between the
returned pose hypotheses and the ground truth. For this purpose, a
dataset containing 90 object poses was used. Specifically, in each
hypotheses set, the pose hypothesis that has the minimum error in
terms of rotation is selected as well as the one with the minimum
translation error. The mean errors for these candidates over the
dataset are shown in Table.~\ref{table:metricEval}. The results
positively indicate the presence of hypotheses close to the true
solution. Specifically, the candidate with the minimum rotation error
seems almost perfect in rotation and not very far even with respect to
translation. Nevertheless, this hypotheses set contained approximately
20,000 elements. It is intractable to evaluate scene level
dependencies for that many hypotheses per object as the combined
hypotheses set over multiple objects grows exponentially in size.

\begin{table*}[h]
    \resizebox{\textwidth}{!}{
    \begin{tabular}{|m{8cm}|m{3.5cm}|m{3.5cm} m{0cm}|}
    \hline
    {\bf Metric for selection} & {\bf Mean Rotation error} & {\bf Mean Translation error} &\\[0.5ex]
    \hline
    [All hypotheses] max. LCP score & 11.16$^{\circ}$ & 1.5cm&\\[0.5ex]
    \hline
    [All hypotheses] min. rotation error from ground truth & 2.11$^{\circ}$ & 2.2cm &\\[0.5ex]
    \hline
    [All hypotheses] min. translation error from ground truth & 16.33$^{\circ}$ & 0.4cm &\\[0.5ex]
    \hline
    [Clustered hypotheses] min. rotation error from ground truth & 5.67$^{\circ}$ & 2.5cm &\\[0.5ex]
    \hline
    [Clustered hypotheses] min. translation error from ground truth & 20.95$^{\circ}$ & 1.7cm&\\[0.5ex]
    \hline
    \end{tabular}}
    \caption{Evaluating the quality of the hypotheses set returned by
      Super4CPS \cite{super4pcs} with respect to different metrics.}
    \label{table:metricEval}
\vspace{-.25in}
\end{table*}

\subsection{Clustering of Hypotheses}

To reduce the cardinality of the hypotheses sets returned by
Algorithm~\ref{alg:alg1}, this work proposes to cluster the 6D poses
in each set {\bf T$_o$} given a distance metric. Computing distances
between object poses, which are defined in SE(3), in a computationally
efficient manner is not trivial \cite{zhang2007c}. This challenge is
further complicated if one would like to consider the symmetry of the
geometric models, so that two different poses that result in the same
occupied volume given the object's symmetry would get a distance of
zero.

To address this issue, a two-level hierarchical clustering approach is
followed. The first level involves computing clusters of the pose set
in the space of translations (i.e., the clustering occurs in
$\mathbb{R}^3$ by using the Euclidean distance and ignoring the object
orientations) using a K-Means process \cite{arthur2007k} to get a
smaller set of cluster representatives {\tt cluster$_{tr}$}. In the
second level, the poses that are assigned to the same clusters are
further clustered based on a distance computed in the SO(3) space that
is specific to the object model, i.e., by considering only the
orientation of the corresponding pose. The second clustering step uses
a {\it kernel} K-Means approach \cite{dhillon2004kernel}, where the
cluster representative is found by minimizing the sum of kernel
distances to every other point in the cluster. This process can be
computationally expensive but returns cluster centers that nicely
represent the accuracy of the hypotheses set. By using this clustering
method, the size of the hypotheses set can be reduced down from 20000
to 25 for each object in our dataset. The two bottom rows of
Table \ref{table:metricEval} evaluate the quality of the cluster
representatives in the hypotheses set. This evaluation indicates that
the clustering process returns hypotheses as cluster representatives
that are still close to the true solution. In this way, it provides an
effective way of reducing the size of the hypotheses set without
sacrificing its diversity.

\subsection{Search}

Once the hypotheses set is built for each object in the scene, the
task reduces to finding the object poses which lie in the physically
consistent neighborhood of the pose candidates that best explain the
overall observed scene. In particular, given:
\begin{itemize}
\item the observed depth image {\tt D}, 
\item the number of objects in the scene {\tt N},
\item a set of 3D mesh models for these objects {\tt M}$_{1:N}$,
\item and the sets of 6D transformation hypotheses for the objects H$_{1:N}$ (output of Algorithm~\ref{alg:alg1}),
\end{itemize}
the problem is to search in the hypotheses sets for an N-tuple of
poses T$_{1:N} \mid T_i \in {\it f}(H_i)$, i.e. one pose per
object. The set $T_{1:N}$ should maximize a global score computed by
comparing the observed depth image {\tt D} with the rendered image
R(T$_{1:N}$) of object models placed at the corresponding poses
T$_{1:N}$. Here, {\it f} is the constrained local optimization of the
object pose $H_I$ based on physical consistency with respect to the
other objects in the scene and also the fact that same points in the
scene point cloud cannot be explained by multiple objects
simultaneously. Then, the global optimization score is defined as:

\vspace{-.1in}
\begin{eqnarray*}
C(D, T_{1:N}) = \sum_{p \in P}S(R(T_{1:N})[p], D[p])
\end{eqnarray*}
where $p$ is a pixel $(i,j)$ of a depth image, $R(T_{1:N})[p]$ is the
depth of pixel $p$ in the rendered depth image $R(T_{1:N})$, $D[p]$ is
the depth of pixel $p$ in the observed depth image {\tt D}, $P
= \{p \mid R(T_{1:N})[p] \neq 0 \text{ or } D[p] \neq 0)\}$ and
\begin{eqnarray*}
S(R(T_{1:N})[p], D[p]) = 
\begin{cases}
1, \text{ if } \mid R(T)[p] - D[p] \mid < \epsilon \cr
0, \text{otherwise}
\end{cases}
\end{eqnarray*}
for a predefined precision threshold $\epsilon$. 
%\begin{align*}
%C(D, T_{1:N}) = \sum_{p \in P}S(R(T_{1:N})[p], D[p])\\
%\text{where }P = \{p \mid R(T_{1:N})[p] \neq 0 \text{ or } D[p] \neq 0)\}\\
%\text{and }S(R(T_{1:N})[p], D[p]) = 
%\begin{cases}
%1, \text{ if } \mid R(T)[p] - D[p] \mid < \epsilon \cr
%0, otherwise
%\end{cases}
%\end{align*}
Therefore, score {\tt C} counts the number of non-zero pixels $p$ that
have a similar depth in the observed image {\tt D} and in the rendered image {\tt R} within an $\epsilon$ threshold. So, overall the objective is to find:
$$T^*_{1:N} = \arg\max_{T_{1:N} \in {\it f}(H_{1:N})} C( D, R(T_{1:N})).$$

At this point a combinatorial optimization problem arises so as to
identify $T^*_{1:N}$, which is approached with a tree search
process. A state in the search-tree corresponds to a subset of objects
in the scene and their corresponding poses. The root state s$_o$ is a
null assignment of poses. A state $s_d$ at depth {\tt d} is a
placement of $d$ objects at specific poses selected from the
hypotheses sets, i.e., {\tt s}$_d = \{{(M_i, T_i),i=1:d}$\} where
$T_i$ is the pose chosen for object $M_i$, which is assigned to depth
$i$. The goal of the tree search is to find a state at depth {\tt N},
which contains a pose assignment for all objects in the scene and
maximizes the above mentioned rendering score. Alg.~\ref{alg:alg2}
describes the expansion of a state in the tree search process towards
this objective.

\vspace{-.1in}
\input{expand}
\vspace{-.1in}

The {\sc expand} routine takes as input, the state {\tt s}$_d$, its
depth $d$ in the tree, maximum tree depth $N$, the point cloud segment
corresponding to the next object to be placed P$_{d+1}$ and the set of
hypotheses $H$. Lines 3-4 of the algorithm iterate over all the
objects already placed in state s$_d$ and remove points explained by
these object placements from the point cloud segment of the next
object to be placed. This step helps in achieving much better
segmentation, which is utilized by the local optimization step of
Trimmed ICP~\cite{chetverikov2002trimmed} in line 5. The poses of
objects according to s$_d$ physically constrain the pose of the new
object to be placed. For this reason, a rigid body physics simulation
is performed in line 6. The physics simulation is initialized by
inserting the new object into the scene at pose $T_{d+1}$, while the
previously inserted objects in the current search branch are placed in
the poses $T_{1:d}$. A physics engine is used to ensure that the newly
placed object attains a physically realistic configuration (stable and
no penetration) with respect to other objects and the table under the
effect of gravity. After a fixed number of simulation steps, the new
pose $T_{d+1}$ of the object is appended to the previous state to get
the successor state s$_{d+1}$.

The above primitive is used to search over the tree of possible object
poses. The objective is to exploit the contextual ordering of object
placements given information from physics and occlusion. This does not
allow to define an additive rendering score over the search depth as
in previous work \cite{Narayanan:2016aa}, which demands the object
placement to not occlude any part of the already placed objects.
Instead, this work proposes to use a heuristic search approach based
on Monte Carlo Tree Search utilizing Upper Confidence Bound
formulation \cite{kocsis2006bandit} to trade off exploration and
exploitation in the expansion process. The pseudocode for the search
is presented in Alg.~\ref{alg:alg3}.

To effectively utilize the constrained expansion of states, an order
of object placements needs to be considered. This information is
encoded in a {\it dependency graph}, which is a directed acyclic graph
that provides a partial ordering of object placements but also encodes
the inderdependency of objects. The vertices of the {\it dependency
graph} correspond to the objects in the observed scene. Simple rules
are established to compute this graph based on the detected segments
{\tt P$_{1:N}$} for objects o$_{1:N}$.
\begin{itemize}
\item A directed edge connects object $o_i$ to object $o_j$ if the the
x-y projection of $P_i$ in the world frame intersects with the x-y
projection of $P_j$ and the z-coordinate (negative gravity direction)
of the centroid for $P_j$ is greater than that of $P_i$.
\item A directed edge connects object $o_i$ to object $o_j$ if the
detected bounding-box of $o_i$ intersects with that of $o_j$ and the
z-coordinate of the centroid of $P_j$ in camera frame (normal to the
camera) is greater than that of $P_i$.
\end{itemize}

The information regarding independency of objects helps to
significantly speed up the search as the independent objects are then
evaluated in different search trees and prevents exponential growth of
the tree. This results in {\tt K} ordered list of objects, {\tt
I}$_{1:K}$, each of which are used to run independent tree searches
for pose computation.  The {\tt MCTS} proceeds by selecting the first
unexpanded node starting from the root state.

\input{search}

%({\sc GET BEST CHILD (s)})

The selection takes place based on a reward associated with each
state. The reward is the mean of the rendering score received at any
leaf node in its subtree along with a penalty based on the number of
times this subtree has been expanded relative to its parent. The
selected state is then expanded by using a {\sc RANDOM\_POLICY}, which
in this case is picking a random object pose hypothesis for each of
the succeeding objects performing the constrained local optimization
at each step. The output of this policy is the final rendering score
of the generated scene hypotheses. This reward is then backpropogated
to the preceeding nodes. Thus, the search is guided to the part of the
tree, which gets a good rendering score but also explores other
portions, which have not been expanded enough (controlled by the
parameter $\alpha$).

%% file: hypogen.tex
\begin{algorithm}[h]
\caption{{\sc Gen\_Hypothesis}( {\tt M}, {\tt cam\_pos}, {\tt RGB}, {\tt D}) }
\label{alg:alg1}
H $\gets$ \{H$_o = \emptyset, \forall o \in$ {\tt M}\}\;
\ForEach{ object o in the scene }{
    bbox$_o \gets$ {\sc RCNN\_detect}( {\tt RGB}, o)\;
    P$_o$ $\gets$ {\sc get\_3DPoints}( bbox$_o$, {\tt D}, {\tt cam\_pos})\;
    T$_o$ $\gets$ {\sc Super4PCS}(M, P$_o$)\;
    \{cluster$_{tr}$, center$_{tr}$\} $\gets$ {\sc KMeans}$_{tr}$(T$_o$)\;
    \ForEach{ cluster c in cluster$_{tr}$ }{
        \{cluster$_{rot}$, center$_{rot}$\} $\gets$ {\sc K-KMeans}$_{tr}$(c)\;
        H$_o$ $\gets$ H$_o$ $\cup$ (center$_{tr}$, center$_{rot}$)\;
    }
    H $\gets$ H $\cup$ H$_o$\;
}
return H;
\end{algorithm}

%% file: expand.tex
\begin{algorithm}[h]
\caption{{\sc expand}$( s_d, T_{d+1}, d, N, P_{d+1} )$ }
\label{alg:alg2}
\If{d $=$ N}{
	return NULL\;
}
\ForEach{ object {\bf o} $\in$ d and {\bf M} }{
	$P_{d+1} \gets P_{d+1}$ - Sim($P_{d+1}, M_oT_o, \epsilon$)\;
}
$T_{d+1} \gets$ TrICP$(T_{d+1}, P_{d+1})$\;
$T_{d+1} \gets$ PHYSIM$(T_{d+1}, T_{1:d})$\;
$s_{d+1} \gets s_d \cup T_{d+1}$\;
return $s_{d+1}$;
\end{algorithm}

%% file: search.tex
\SetKwProg{Fn}{Function}{}{}

\begin{algorithm}[h]
\caption{{\sc search} }
\label{alg:alg3}
\Fn{COMPUTE\_POSE (M$_{1:N}$, N, P$_{1:N}$, H$_{1:N}$)}{
	Res $\gets \emptyset$\;
	$I_{1:K} \gets$ {\sc get\_dependency}$(M_{1:N}, P_{1:N}, H_{1:N})$\;
	\ForEach{ordered list L $\in I_{1:K}$}{
		Res $\gets$ Res $\cup$ MCTS($L_o, N, P_{1:N}, H_{1:N}$)\;
	}
	return Res\;
}
\BlankLine
\Fn{MCTS ($s_o, N, P_{1:N}, H_{1:N}$)}{
	\While{$search\_time < t_{th}$}{
		$s_i \gets$ {\sc Select}$(s_o, N, P_{1:N})$\;
		R $\gets$ {\sc random\_policy}($s_i, N, P_{1:N}$)\;
		{\sc backup\_reward}($s_i, R$)\;
	}
}
\BlankLine
\Fn{SELECT ($s, N, P_{1:N}$)}{
	\While{depth($s$) $<$ N}{
		\eIf{$s$ has unexpanded child}{
			oIdx $\gets$ $depth(s) + 1$\;
			T$_{oIdx}$ $\gets$ {\sc get\_pose\_hypothesis}(H$_{oIdx}$)\;
			return {\sc expand}($s, T_{oIdx}, depth(s), N, P_{oIdx}$)\;
		}{
			s $\gets$ {\sc get\_best\_child}(s)
		}
	}
	return s\;
}
\BlankLine
\Fn{GET\_BEST\_CHILD ($s$)}{
	return arg max$_{s' \in succ(s)} \frac{h(s')}{n(s')} + \alpha \sqrt{\frac{2ln n(s)}{n(s')}}$\;
}
\BlankLine
\Fn{RANDOM\_POLICY ($s, N, P_{1:N}$)}{
	\While{depth($s$) $<$ N}{
		oIdx $\gets$ $depth(s) + 1$\;
		T$_{oIdx}$ $\gets$ {\sc get\_random\_hypothesis}(H$_{oIdx}$)\;
		s $\gets$ {\sc expand}($s, T_{oIdx}, depth(s), N, P_{oIdx}$)\;
	}
	return render(s)\;
}
\BlankLine
\Fn{BACKUP\_REWARD ($s, R$)}{
	\While{s $\neq$ NULL}{
		n(s) $\gets$ n(s) + 1\;
		h(s) $\gets$ h(s) + R\;
		s $\gets$ parent(s)\;
	}
}
\end{algorithm}

%% file: 04_evaluation.tex
This section discusses the dataset and metrics used for evaluating the
approach, and an analysis of intermediate results explaining the
choice of the system's components.

\subsection{Dataset and Evaluation Metric}

A dataset of RGB-D images was collected and ground truth {\tt 6-DOF}
poses were labeled for each object in the image. The dataset contains
table-top scenes with 11 objects from the 2016 Amazon Picking
Challenge \cite{APC2016} with the objects representing different
object geometries. Each scene contains multiple objects and the object
placement is a mix of independent object placements, objects with
physical dependencies such as one stacked on/or supporting the other
object and occlusions. The dataset was collected using an Intel
RealSense sensor mounted over a Motoman robotic manipulator. The
manual labeling was achieved by aligning 3D CAD models to the point
cloud extracted from the sensor. The captured scene expresses three
different levels of interaction between objects, namely, independent
object placement where an object is physically independent of the rest
of objects, two-object dependencies where an object depends on
another, and three object dependencies where an object depends on two
other objects.

The evaluation is performed by computing the error in translation,
which is the Euclidean distance of an object's center compared to its
ground truth center (in centimeters). The error in rotation is
computed by first transforming the computed rotation to the frame
attached to the object at ground truth (in degrees). The rotation
error is the average of the roll, yaw and pitch angles of the
transformation between the returned rotation and the ground truth one,
while taking into account the object's symmetries, which may allow
multiple ground truth rotations. The results provide the mean of the
errors of all the objects in the dataset.

\begin{table*}[ht]
\resizebox{\textwidth}{!}{
\begin{tabular}{|m{5cm}|c c|c c|c c|c c m{0cm}|}
\hline
\centering{\bf Method} & \multicolumn{2}{c|}{One-object Dependencies} & \multicolumn{2}{c|}{Two-objects Dependencies} & \multicolumn{2}{c|}{Three-objects Dependencies} & \multicolumn{2}{c}{All} &\\[0.5ex]
\hline
& Rotation Error & Translation Error & Rotation Error & Translation Error & Rotation Error & Translation Error&Rotation Error & Translation Error &\\[0.5ex]
\hline
APC-Vision-Toolbox~\cite{Princeton} & 15.5$^{\circ}$ & 3.4 cm & 26.3$^{\circ}$ & 5.5 cm& 17.5$^{\circ}$ & 5.0 cm& 21.2$^{\circ}$ & 4.8 cm&\\[0.5ex]
\hline
faster-RCNN + PCA + ICP~\cite{Mitash:2017aa} & 8.4$^{\circ}$ & 1.3 cm& 13.1$^{\circ}$ & 2.0 cm& 12.3$^{\circ}$ & 1.8 cm& 11.6$^{\circ}$ & 1.7 cm&\\[0.5ex]
\hline
faster-RCNN + Super4PCS + ICP~\cite{Hernandez:2016aa, Mitash:2017aa} & 2.4$^{\circ}$ & 0.8 cm& 14.8$^{\circ}$ & 1.7 cm& 12.1$^{\circ}$ & 2.1 cm& 10.5$^{\circ}$ & 1.5 cm&\\[0.5ex]
\hline
PHYSIM-Heuristic (depth + LCP) & 2.8$^{\circ}$ & 1.1 cm& 5.8$^{\circ}$ & 1.4 cm& 12.5$^{\circ}$ & 3.1 cm& 6.3$^{\circ}$ & 1.7 cm&\\[0.5ex]
\hline
PHYSIM-MCTS (proposed approach) & 2.3$^{\circ}$ & 1.1 cm& 5.8$^{\circ}$ & 1.2 cm& 5.0$^{\circ}$ & 1.8 cm& 4.6$^{\circ}$ & 1.3 cm &\\[0.5ex]
\hline
\end{tabular}}
\caption{Comparing our approach with different pose estimation techniques}% based on segmentation followed by point cloud registration}
\label{table:poseEst}
\vspace{-.25in}
\end{table*}

% \hline
% MCTS (Without Physics) & 8.06 & 1.0 &\\[1ex]

\subsection{Pose Estimation without Search}

Evaluation was first performed over methods that do not perform any
scene level or global reasoning. These approaches trust the segments
returned by the object segmentation module and perform model matching
followed by local refinement to compute object poses. The results of
performing pose estimation over the collected dataset with some of
these techniques are presented in Table~\ref{table:poseEst}. The {\tt
  APC-Vision-Toolbox} \cite{Princeton} is the system developed by Team
MIT-Princeton for Amazon Picking Challenge 2016. The system uses a
Fully Convolutional Network ({\tt FCN}) \cite{shelhamer2016fully} to
get pixel level segmentation of objects in the scene, then uses
Principal Component Analysis ({\tt PCA}) for pose initialization,
followed by {\tt ICP} \cite{icp} to get the final object pose. This
system was designed for shelf and tote environments and often relies
on multiple views of the scene. Thus, the high error in pose estimates
could be attributed to the low recall percentage in retrieving object
segment achieved by the semantic segmentation method, which in turn
resulted in the segment not having enough information to compute a
unique pose estimate. The second system tested uses a {\tt
  Faster-RCNN}-based object detector trained with a setup-specific and
autonomously generated dataset \cite{Mitash:2017aa}. The point cloud
segments extracted from the bounding box detections were used to
perform pose estimation using two different approaches: i) {\tt PCA}
followed by {\tt ICP} and ii) {\tt Super4PCS} followed by {\tt ICP}
\cite{icp}. Even though the detector succeeded in providing a high
recall object segment on most occasions, in the best case the mean
rotation error using local approaches was still high
({10.5}$^{\circ}$). This was sometimes due to bounding boxes
containing parts of other object segments, or due to
occlusions. Reasoning only at a local object-level does not resolve
these issues.

\begin{figure}[h]
  \vspace{-.15in}
    \centering \includegraphics[width=\linewidth,
      keepaspectratio]{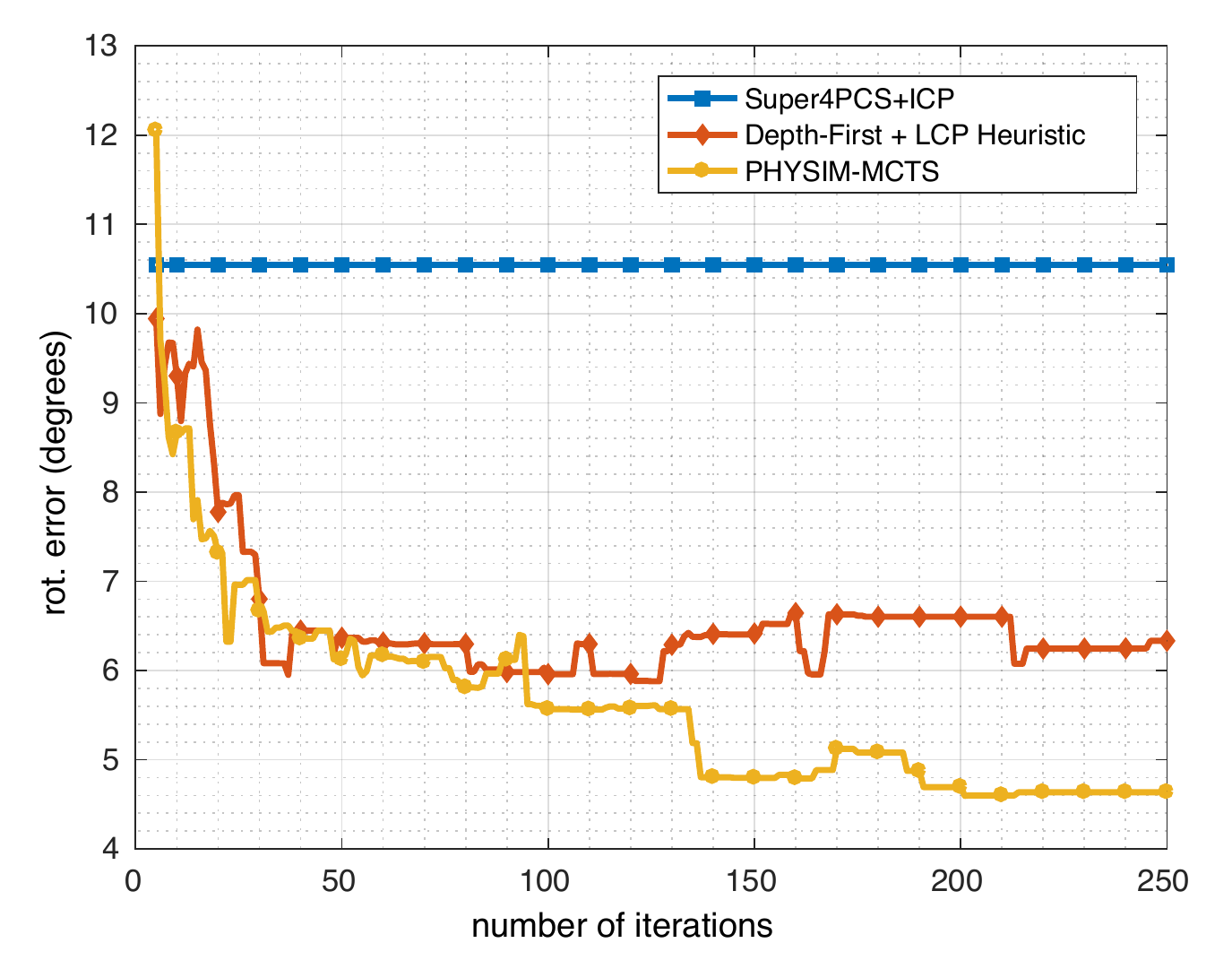}
      \vspace{-.3in}
    \caption{Rotation error in degrees as a function of the number of
      iterations.}
      \vspace{-.2in}

    \label{fig:rotErr}
\end{figure}

\begin{figure}[h]    
  \vspace{-.15in}
    \centering \includegraphics[width=\linewidth,
      keepaspectratio]{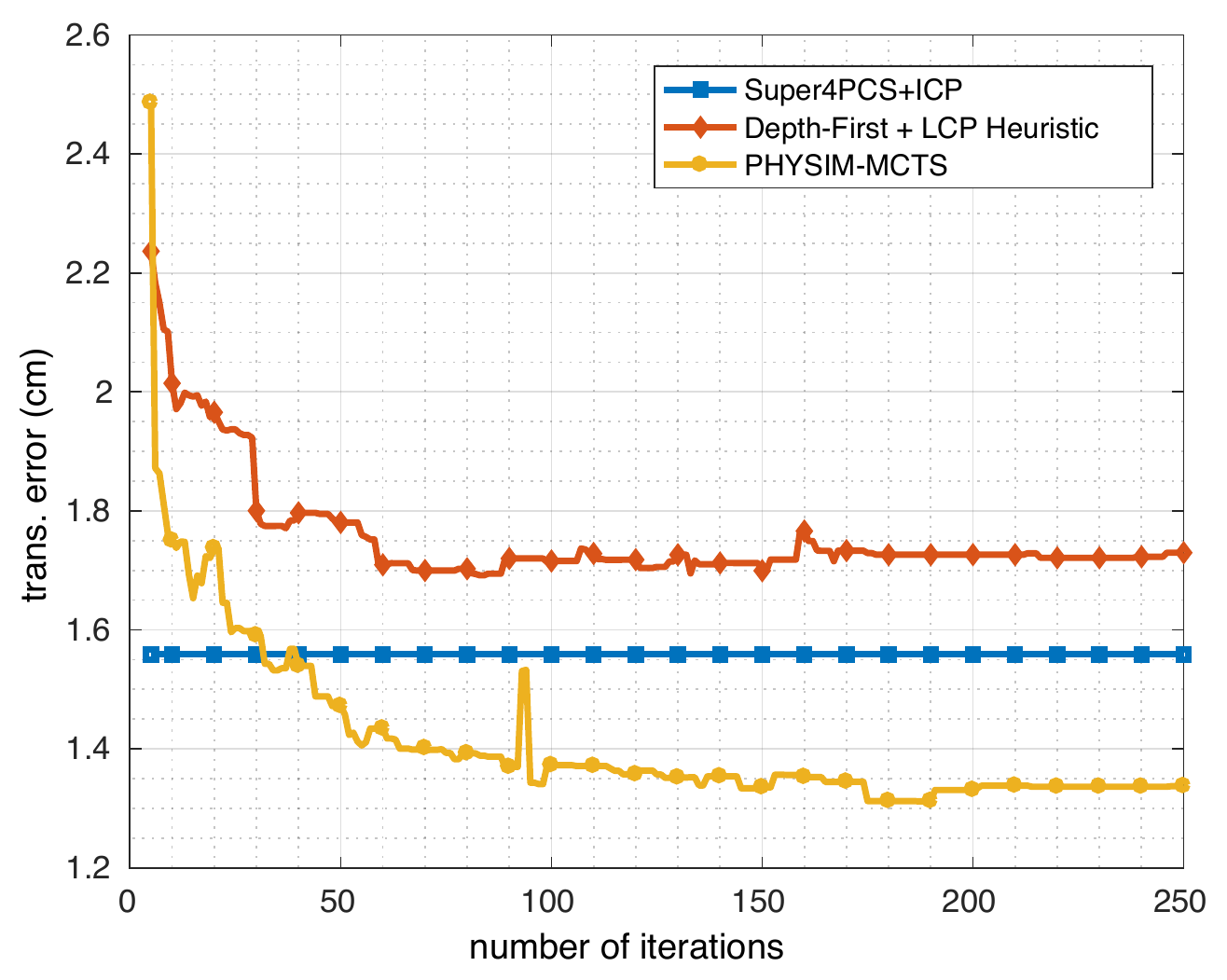}    
  \vspace{-.3in}
    \caption{Translation error in cm as a function of the number of iterations.}
    \label{fig:transErr}
  \vspace{-.15in}
\end{figure}

\subsection{Pose Estimation with the proposed approach}

The proposed search framework was used to perform pose estimation on
the dataset. In each scene, the dependency graph structure was used to
get the order of object placement and initialize the independent
search trees. Then, the object detection was performed using {\tt
  Faster-RCNN} and {\tt Super4PCS} was used to generate pose
candidates, which were clustered to get 25 representatives per
object. The search is performed over the combined set of object
candidates and the output of the search is an anytime pose estimate
based on the best rendering score. The stopping criterion for the
searches was defined by a maximum number of node expansions in the
tree, set to 250, where each expansion corresponds to a physics
simulation with {\it Bullet} and a rendering with {\it OpenGL}, with a
mean expansion time of $\sim0.2$ secs per node. The search was
initially performed using a depth-first heuristic combined with the
LCP score returned by the Super4PCS for the pose candidates. The
results from this approach, {\tt PHYSIM-Heuristic (depth + LCP)}, are
shown in Table~\ref{table:poseEst}, which indicates that it might be
useful to use these heuristics if the tree depth is low (one and two
object dependencies). As the number of object dependencies grow,
however, one needs to perform more exploration. For three-object
dependencies, when using 250 expansions, this heuristic search
provided poor performance. The UCT Monte Carlo Tree Search was used to
perform the search, with upper confidence bounds to trade off
exploration and exploitation.  The exploration parameter was set to a
high value ($\alpha = 5000$), to allow the search to initially look
into more branches while still preferring the ones which give a high
rendering score. This helped in speeding up the search process
significantly, and a much better solution could be reached within the
same time. The plots in Fig.~\ref{fig:rotErr} and
Fig.~\ref{fig:transErr} captures the anytime results from the two
heuristic search approaches.

\subsection{Limitations}

One of the limitations of global reasoning as is in this approach is
the large amount of time required for computing and searching over an
extensive hypotheses set. Particularly, due to the hierarchical
clustering approach, which was adapted to consider object specific
distances, the hypotheses generation time for an object can be in the
order of multiple seconds. The search process, which seemed to
converge to good solutions with 150 expansions for three-object
dependencies, takes approximately 30 seconds. Nevertheless, both of
these processes are highly parallelizable. Future work can perform the
hypotheses generation and the search with parallel computing.  Even
though the use of bounding boxes to compute 3D point segments gives a
high recall in terms of obtaining the object point segment, and the
proposed system also addresses any imprecision which might occur in
these segments by performing a constrained segmentation, sometimes an
error which occurs in the initial part of object placement could lead
to failures that need to be addressed.

%% file: 05_conclusion.tex
This work provides a novel way of performing pose estimation for
objects placed in clutter by efficiently searching for the best scene
explanation over the space of physically consistent scene
configurations. It also provides a method to construct these sets of
scene configurations by using state-of-the-art object detection and
model registration techniques which by themselves are not sufficient
to give a desirable pose estimate for objects.  The evaluations
indicate significant performance improvement in the estimated pose
using the proposed search, when compared to systems which do not
reason globally. Moreover, the use of Monte Carlo Tree search with the
scene rendering cost evaluated over physically simulated scenes makes
the search tractable for handling multi-object dependencies.